\pdfoutput=1

\documentclass{article}
\usepackage{colm2024_conference}

\usepackage{url}
\usepackage{booktabs}
\usepackage{hyperref}
\usepackage{graphicx}
\usepackage{microtype}
\usepackage{multicol}
\usepackage{xcolor}   
\usepackage{multirow}
\usepackage{lipsum}
\usepackage{enumitem}
\usepackage{pifont}
\usepackage{colortbl}
\usepackage{tabularx}
\usepackage{caption} 
\usepackage{float}
\usepackage{amsmath,amsfonts,amssymb,bbm}

\newcommand*\samethanks[1][\value{footnote}]{\footnotemark[#1]}
\newcommand{\squad}{\quad\ }

\title{Group Sequence Policy Optimization}

\author{
\textbf{Chujie Zheng\thanks{Corresponding authors.} \squad Shixuan Liu \squad Mingze Li \squad Xiong-Hui Chen \squad Bowen Yu\samethanks{} \\ Chang Gao \squad Kai Dang  \squad Yuqiong Liu \squad Rui Men \squad An Yang \squad Jingren Zhou \squad Junyang Lin } \\
\vspace{1.5mm}
Qwen Team, Alibaba Inc.
}

\begin{document}
\maketitle

\begin{abstract}

This paper introduces Group Sequence Policy Optimization (GSPO), our stable, efficient, and performant reinforcement learning algorithm for training large language models.
Unlike previous algorithms that adopt token-level importance ratios, GSPO defines the importance ratio based on sequence likelihood and performs sequence-level clipping, rewarding, and optimization.
We demonstrate that GSPO achieves superior training efficiency and performance compared to the GRPO algorithm, notably stabilizes Mixture-of-Experts (MoE) RL training, and has the potential for simplifying the design of RL infrastructure.
These merits of GSPO have contributed to the remarkable improvements in the latest Qwen3 models.

\end{abstract}

\section{Introduction}
\label{sec:intro}

Reinforcement learning (RL) has emerged as a pivotal paradigm for scaling language models \citep{o1,dpsk-r1,qwq32b,qwen3}.
Through large-scale RL, language models develop the capability to tackle sophisticated problems, such as competition-level mathematics and programming, by undertaking deeper and longer reasoning processes.

To successfully scale RL with greater computational investment, the foremost prerequisite is maintaining stable and robust training dynamics.
However, current state-of-the-art RL algorithms, exemplified by GRPO \citep{grpo}, exhibit severe stability issues when training gigantic language models, often resulting in catastrophic and irreversible model collapse \citep{qwen3, minimax-m1}.
This instability hinders efforts to push the capability boundaries of language models through continued RL training.

In this paper, we identify that the instability of GRPO stems from the fundamental misapplication and invalidation of importance sampling weights in its algorithmic design. 
This introduces high-variance training noise that progressively accumulates with increased response length and is further amplified by the clipping mechanism, ultimately precipitating model collapse.

To address these core limitations, we propose \textbf{Group Sequence Policy Optimization (GSPO)}, a new RL algorithm for training large language models.
The key innovation of GSPO lies in its theoretically grounded definition of importance ratio based on sequence likelihood \citep{click}, aligning with the basic principle of importance sampling.
Additionally, GSPO computes the normalized rewards as the advantages of multiple responses to a query, ensuring the alignment between sequence-level rewarding and optimization.

Our empirical evaluation demonstrates the significant superiority of GSPO over GRPO in training stability, efficiency, and performance.
Critically, GSPO has inherently resolved the stability challenges in the RL training of large Mixture-of-Experts (MoE) models, eliminating the need for complex stabilization strategies, and shows the potential for simplifying RL infrastructure.
These merits of GSPO ultimately contributed to the exceptional performance improvements in the latest Qwen3 models.
We envision GSPO as a robust and scalable algorithmic foundation that will enable the continued advancement of large-scale RL training with language models.

\section{Preliminaries}

\paragraph{Notation}
In this paper, an autoregressive language model parameterized by $\theta$ is defined as a policy $\pi_\theta$.
We use $x$ to denote a query and $\mathcal{D}$ as the query set.
Given a response $y$ to a query $x$, its likelihood under the policy $\pi_\theta$ is denoted as $\pi_\theta (y | x)=\prod_{t=1}^{|y|} \pi_\theta (y_t | x, y_{<t} )$ where $|y|$ denotes the number of tokens in $y$.
A query-response pair $(x, y)$ can be scored by a verifier $r$, resulting in a reward $r(x, y) \in [ 0, 1 ]$.

\paragraph{Proximal Policy Optimization (PPO)}
Using samples generated from the old policy $\pi_{\theta_\text{old}}$, PPO \citep{ppo} constrains the policy update within a proximal region of the old policy through the clipping mechanism.
Specifically, PPO employs the following objective for policy optimization (we omit the KL regularization term hereinafter for brevity, as it is not the focus of this paper):
\begin{align}
\mathcal{J}_\text{PPO}(\theta) = \mathbb{E}_{ x \sim \mathcal{D},\, y \sim \pi_{\theta_\text{old}}( \cdot | x) }
\left[ \frac{1}{|y|} \sum_{t=1}^{|y|} 
\min \left( w_{t}(\theta) \widehat{A}_{t},  \, \mathrm{clip} \left( w_{t}(\theta), 1 - {\varepsilon}, 1 + {\varepsilon}\right) \widehat{A}_{t} \right)
\right],
\end{align}
where the importance ratio of the token $y_t$ is defined as
$
w_{t}(\theta) = \frac{ \pi_{\theta} (y_{t} | x, y_{<t}) }{ \pi_{\theta_\text{old}} (y_{t} | x,y_{<t})}
$,
the advantage $\widehat{A}_{t}$ of $y_t$ is estimated by another value model, and $\varepsilon$ is the clipping range of importance ratios.

The core challenge of PPO in practice lies in its heavy reliance on the value model.
Specifically, the value model usually has a similar size to the policy model, introducing a considerable memory and computational burden.
Furthermore, the algorithmic effectiveness hinges on the reliability of its value estimate.
While acquiring a reliable value model is inherently challenging, ensuring its scalability to longer responses and more complex tasks presents an even greater challenge.

\paragraph{Group Relative Policy Optimization (GRPO)}
GRPO \citep{grpo} bypasses the need for the value model by computing the relative advantage of each response within a group of responses to the same query.
Specifically, GRPO optimizes the following objective:
\begin{align}
\label{equ:grpo}
\mathcal{J}_\text{GRPO}(\theta) = \mathbb{E}_{ x \sim \mathcal{D},\, \{y_i\}_{i=1}^G \sim \pi_{\theta_\text{old}}( \cdot | x) }
\left[ \frac{1}{G} \sum_{i=1}^{G} \frac{1}{|y_i|} \sum_{t=1}^{|y_i|} 
\min \left( w_{i,t}(\theta) \widehat{A}_{i,t},  \, \mathrm{clip} \left( w_{i,t}(\theta), 1 - {\varepsilon}, 1 + {\varepsilon}\right) \widehat{A}_{i,t} \right)
\right],
\end{align}
where $G$ is the number of generated responses to each query $x$ (i.e., the group size), and the importance ratio $w_{i,t}(\theta)$ and advantage $\widehat{A}_{i,t}$ of token $y_{i,t}$ are:
\begin{align}
    w_{i,t}(\theta)=\frac{ \pi_{\theta} (y_{i,t} | x, y_{i,<t}) }{ \pi_{\theta_\text{old}} (y_{i,t} | x,y_{i,<t})},\squad
    \widehat{A}_{i,t} = \widehat{A}_{i} = \frac{r(x, y_i) - \mathrm{mean} \left( \{ r(x, y_i) \}_{i=1}^G \right) }{ \mathrm{std} \left( \{ r(x, y_i) \}_{i=1}^G \right) },
\end{align}
respectively, where all the tokens in $y_i$ share the same advantage as $\widehat{A}_{i}$.

\section{Motivation}
\label{sec:motivation}

The growth in model size, sparsity (e.g., in Mixture-of-Experts models), and response length necessitates a large rollout batch size to maximize hardware utilization during RL.
To improve sample efficiency, it is standard practice to partition a large batch of rollout data into multiple mini-batches for gradient updates.
This procedure inevitably introduces an off-policy learning setting, where responses $y$ are sampled from an old policy $\pi_{\theta_\text{old}}$ rather than the current policy $\pi_\theta$ being optimized.
This also explains the necessity of the clipping mechanism in PPO and GRPO, which prevents overly ``off-policy'' samples from being involved in gradient estimation.

While mechanisms like clipping aim to manage this off-policy discrepancy, we identify a more fundamental issue in GRPO: \textit{its objective is ill-posed}.
This problem becomes particularly acute when training large models on long-response tasks, leading to catastrophic model collapse.
The ill-posed nature of the GRPO objective stems from a misapplication of importance sampling weights.
The principle of importance sampling is to estimate the expectation of a function $f$ under a target distribution $\pi_\text{tar}$ by re-weighting samples drawn from a behavior distribution $\pi_\text{beh}$:
\begin{align}
\mathbb{E}_{z \sim \pi_\text{tar}} \left[ f(z) \right] = \mathbb{E}_{z \sim \pi_\text{beh}} \left[ \frac{ \pi_\text{tar} (z) }{ \pi_\text{beh} (z)} \, f(z) \right].
\end{align}
Crucially, this relies on averaging over multiple samples ($N \gg 1$) from the behavior distribution $\pi_\text{beh}$ for the importance weight $\frac{ \pi_\text{tar} (z) }{ \pi_\text{beh} (z)}$ to effectively correct for the distributional mismatch.

In contrast, GRPO applies the importance weight $\frac{ \pi_{\theta} (y_{i,t} | x, y_{i,<t}) }{ \pi_{\theta_\text{old}} (y_{i,t} | x,y_{i,<t})}$ at each token position $t$.
Since this weight is based on a single sample $y_{i,t}$ from each next-token distribution $\pi_{\theta_\text{old}}( \cdot | x, y_{i, <t})$, it fails to perform the intended distribution-correction role.
Instead, it introduces high-variance noise into the training gradients, which accumulates over long sequences and is exacerbated by the clipping mechanism.
We have empirically observed that this can lead to model collapse that is often irreversible.
Once the collapse occurs, resuming training is unavailing, even when reverting to a previous checkpoint and meticulously tuning hyperparameters (e.g., the clipping ranges), extending generation length, or switching the RL queries.

The above observation suggests a fundamental issue in GRPO's design.
The failure of the token-level importance weight points to a core principle: \textbf{the unit of optimization objective should match the unit of reward}.
Since the reward is granted to the entire sequence, applying off-policy correction at the token level appears problematic.
This motivates us to forego the token-level objective and explore utilizing importance weights and performing optimization directly at the \textit{sequence level}.

\section{Algorithm}

\subsection{GSPO: Group Sequence Policy Optimization}

While the token-level importance weight $\frac{ \pi_{\theta} (y_{i,t} | x, y_{i,<t}) }{ \pi_{\theta_\text{old}} (y_{i,t} | x,y_{i,<t})}$ is problematic in GRPO, we observe that in the context of language generation, the \textit{sequence-level} importance weight $\frac{ \pi_{\theta} (y | x) }{ \pi_{\theta_\text{old}} (y | x)}$ has a clear theoretical meaning: it reflects how far the response $y$ sampled from $\pi_{\theta_\text{old}} (\cdot | x)$ deviates from $\pi_{\theta} (\cdot | x)$, which naturally aligns with the sequence-level reward and can also serve as a meaningful indicator of the clipping mechanism.

\vspace{1.5ex}

Based on this straightforward observation, we propose the \textbf{Group Sequence Policy Optimization (GSPO)} algorithm.
GSPO employs the following sequence-level optimization objective:
\begin{align}
\mathcal{J}_\text{GSPO} (\theta) =
\mathbb{E}_{ x \sim \mathcal{D},\, \{y_i\}_{i=1}^G \sim \pi_{\theta_\text{old}}( \cdot | x) }
\left[ 
\frac{1}{G} \sum_{i=1}^{G}
\min \left( s_{i}(\theta)  \widehat{A}_{i},  \, \mathrm{clip} \left( s_{i}(\theta), 1 - {\varepsilon}, 1 + {\varepsilon} \right) \widehat{A}_{i} \right) 
\right],
\label{equ:gspo}
\end{align}
where we adopt the group-based advantage estimation:
\begin{align}
\widehat{A}_{i} = \frac{r(x, y_i) - \mathrm{mean} \left( \{ r(x, y_i) \}_{i=1}^G \right) }{ \mathrm{std} \left( \{ r(x, y_i) \}_{i=1}^G \right) },
\end{align}
and define the importance ratio $s_{i}(\theta)$ based on sequence likelihood \citep{click}:
\begin{align}
s_{i}(\theta) = \left( \frac{ \pi_{\theta} (y_i | x) }{ \pi_{\theta_\text{old}} (y_i | x)} \right)^{\frac{1}{|y_i|}}
=
\exp \left( \frac{1}{|y_i|} \sum_{t=1}^{|y_i|} \log \frac{ \pi_{\theta} (y_{i,t} | x, y_{i,<t}) }{ \pi_{\theta_\text{old}} (y_{i,t} | x,y_{i,<t})} \right).
\end{align}
Therefore, GSPO applies clipping to entire responses instead of individual tokens to exclude the overly ``off-policy'' samples from gradient estimation, which matches both the sequence-level rewarding and optimization.
Note that we adopt length normalization in $s_{i}(\theta)$ to reduce the variance and to control  $s_{i}(\theta)$ within a unified numerical range.
Otherwise, the likelihood changes of a few tokens can result in dramatic fluctuations of the sequence-level importance ratio, and the importance ratios of responses with different lengths will require varying clipping ranges.
We also note that the clipping ranges in GSPO and in previous algorithms (e.g., GRPO) typically differ in order of magnitude due to the distinct definitions of importance ratios.

\subsection{Gradient Analysis}
\label{subsec:gradient}

We can derive the gradient of the GSPO objective as follows (clipping is omitted for brevity):
\begin{align}
\nabla_{\theta} \mathcal{J}_\text{GSPO} (\theta)
=&\ 
\nabla_{\theta} \mathbb{E}_{ x \sim \mathcal{D},\, \{y_i\}_{i=1}^G \sim \pi_{\theta_\text{old}}( \cdot | x) }
\left[ 
\frac{1}{G} \sum_{i=1}^{G} s_{i}(\theta) \widehat{A}_{i}
\right] \\
= &\ 
\mathbb{E}_{ x \sim \mathcal{D},\, \{y_i\}_{i=1}^G \sim \pi_{\theta_\text{old}}( \cdot | x) }
\left[ 
\frac{1}{G} \sum_{i=1}^{G}
s_{i}(\theta) \widehat{A}_{i}
\cdot \nabla_{\theta} \log s_{i}(\theta)
\right] \\
=&\ 
\mathbb{E}_{ x \sim \mathcal{D},\, \{y_i\}_{i=1}^G \sim \pi_{\theta_\text{old}}( \cdot | x) }
\left[ 
\frac{1}{G} \sum_{i=1}^{G} 
\left( \frac{ \pi_{\theta} (y_i | x) }{ \pi_{\theta_\text{old}} (y_i | x)} \right)^{\frac{1}{|y_i|}} \widehat{A}_{i} 
\cdot \frac{1}{|y_i|} \sum_{t=1}^{|y_i|} \nabla_{\theta} \log \pi_{\theta} (y_{i,t} | x, y_{i,<t}) 
\right].
\label{equ:gspo_grad}
\end{align}
For comparison, the gradient of the GRPO objective is as follows (note that $\widehat{A}_{i,t} = \widehat{A}_{i}$):
\begin{align}
\nabla_{\theta} \mathcal{J}_\text{GRPO} (\theta) 
=&\ 
\nabla_{\theta} \mathbb{E}_{ x \sim \mathcal{D},\, \{y_i\}_{i=1}^G \sim \pi_{\theta_\text{old}}( \cdot | x) }
\left[ \frac{1}{G} \sum_{i=1}^{G} \frac{1}{|y_i|} \sum_{t=1}^{|y_i|} 
w_{i,t}(\theta) \widehat{A}_{i,t}
\right] \\
=&\
\mathbb{E}_{ x \sim \mathcal{D},\, \{y_i\}_{i=1}^G \sim \pi_{\theta_\text{old}}( \cdot | x) }
\left[ \frac{1}{G} \sum_{i=1}^{G} \widehat{A}_{i} 
\cdot \frac{1}{|y_i|} \sum_{t=1}^{|y_i|} 
\frac{ \pi_{\theta} (y_{i,t} | x, y_{i,<t}) }{ \pi_{\theta_\text{old}} (y_{i,t} | x,y_{i,<t})} 
\nabla_{\theta} \log \pi_{\theta} (y_{i,t} | x, y_{i,<t})  
\right].
\end{align}
Therefore, the fundamental distinction between GSPO and GRPO lies in \textit{how they weight the gradients of the log likelihoods of tokens}.
In GRPO, the tokens are weighted according to their respective ``importance weight'' $\frac{ \pi_{\theta} (y_{i,t} | x, y_{i,<t}) }{ \pi_{\theta_\text{old}} (y_{i,t} | x,y_{i,<t})}$.
However, these unequal weights, which can vary among $(0, 1+\varepsilon]$ (for $\widehat{A}_{i} >0$) or $[1-\varepsilon, +\infty)$ (for $\widehat{A}_{i} < 0$), are not negligible, and their impact can accumulate and lead to unpredictable consequences as training progresses.
In contrast, GSPO weights all the tokens in a response equally, eliminating this instability factor of GRPO.

\subsection{GSPO-token: A Token-level Objective Variant}

In scenarios like multi-turn RL, we may desire a finer-grained advantage adjustment than the sequence level.
To this end, we introduce a token-level objective variant of GSPO, namely \textbf{GSPO-token}, to allow token-wise advantage customization:
\begin{align}
\mathcal{J}_\text{GSPO-token}(\theta)
=
\mathbb{E}_{ x \sim \mathcal{D},\, \{y_i\}_{i=1}^G \sim \pi_{\theta_\text{old}}( \cdot | x) }
\left[ 
\frac{1}{G} \sum_{i=1}^{G} \frac{1}{|y_i|} \sum_{t=1}^{|y_i|} 
\min \left( s_{i,t}(\theta) \widehat{A}_{i,t},  \, \mathrm{clip} \left( s_{i,t}(\theta), 1 - {\varepsilon}, 1 + {\varepsilon} \right) \widehat{A}_{i,t} \right) 
\right],
\label{equ:gspo-token}
\end{align}
where
\begin{align}
s_{i,t}(\theta) 
= \mathrm{sg} \left[ s_{i}(\theta) \right]  \cdot \frac{ \pi_{\theta} (y_{i,t} | x, y_{i,<t}) }{ \mathrm{sg} \left[ \pi_{\theta} (y_{i,t} | x, y_{i,<t}) \right] },
\end{align}
and $\mathrm{sg}[\cdot]$ denotes only taking the numerical value but stopping the gradient, corresponding to the \texttt{detach} operation in PyTorch.
The gradient of GSPO-token can be derived as:
\begin{align}
\nabla_{\theta} \mathcal{J}_\text{GSPO-token}(\theta)
=&\ 
\nabla_{\theta} \mathbb{E}_{ x \sim \mathcal{D},\, \{y_i\}_{i=1}^G \sim \pi_{\theta_\text{old}}( \cdot | x) }
\left[ 
\frac{1}{G} \sum_{i=1}^{G}
\frac{1}{|y_i|} \sum_{t=1}^{|y_i|} 
s_{i,t}(\theta) \widehat{A}_{i,t}
\right] \\
=&\ 
\mathbb{E}_{ x \sim \mathcal{D},\, \{y_i\}_{i=1}^G \sim \pi_{\theta_\text{old}}( \cdot | x) }
\left[ 
\frac{1}{G} \sum_{i=1}^{G} s_i (\theta)
\cdot \frac{1}{|y_i|} \sum_{t=1}^{|y_i|} 
\widehat{A}_{i,t} \frac{ \nabla_{\theta} \pi_{\theta} (y_{i,t} | x, y_{i,<t}) }{ \pi_{\theta} (y_{i,t} | x, y_{i,<t}) }
\right] \\
=&\ 
\mathbb{E}_{ x \sim \mathcal{D},\, \{y_i\}_{i=1}^G \sim \pi_{\theta_\text{old}}( \cdot | x) }
\left[ 
\frac{1}{G} \sum_{i=1}^{G}  \left( \frac{ \pi_{\theta} (y_i | x) }{ \pi_{\theta_\text{old}} (y_i | x)} \right)^{\frac{1}{|y_i|}}
\cdot \frac{1}{|y_i|} \sum_{t=1}^{|y_i|}
\widehat{A}_{i,t} \nabla_{\theta} \log \pi_{\theta} (y_{i,t} | x, y_{i,<t})  
\right].
\label{equ:gspo-token_grad}
\end{align}
Note that the term $\frac{ \pi_{\theta} (y_{i,t} | x, y_{i,<t}) }{ \mathrm{sg} \left[ \pi_{\theta} (y_{i,t} | x, y_{i,<t}) \right] }$ has a numerical value of 1, so $s_{i,t}(\theta)$ is numerically equal to $s_{i}(\theta)$.
Comparing Equation~\eqref{equ:gspo} and~\eqref{equ:gspo-token}, and Equation~\eqref{equ:gspo_grad} and~\eqref{equ:gspo-token_grad}, GSPO-token and GSPO are numerically identical in the optimization objective, clipping condition, and theoretical gradient when we set the advantages of all the tokens in the response $y_i$ to the same value (i.e., $\widehat{A}_{i,t} = \widehat{A}_{i}$), while GSPO-token enjoys the higher flexibility of adjusting the advantages per token.

\section{Experiments and Discussion}

\subsection{Empirical Results}

We experiment with a cold-start model fine-tuned from Qwen3-30B-A3B-Base, and report the training reward curves as well as the model performance curves on the AIME'24 (average Pass@1 over 32 samplings), LiveCodeBench (202410-202502, average Pass@1 over 8 samplings), and CodeForces (Elo Rating) benchmarks.
During the RL training, each batch of rollout data is partitioned into four mini-batches for gradient updates.
In GSPO, we set the left and right clipping ranges in Equation~\eqref{equ:gspo} to 3e-4 and 4e-4, respectively.
We compare against GRPO as the baseline and set the left and right clipping ranges in Equation~\eqref{equ:grpo} to 0.2 and 0.27, respectively, which we have carefully tuned to ensure a fair comparison.
Note that GRPO necessitates the Routing Replay training strategy for the normal convergence of MoE RL, which we will additionally discuss in \S~\ref{subsec:routing_replay}, while \textit{GSPO has obviated the need for this strategy}.

\begin{figure}[htbp]
    \centering
    \includegraphics[width=\linewidth]{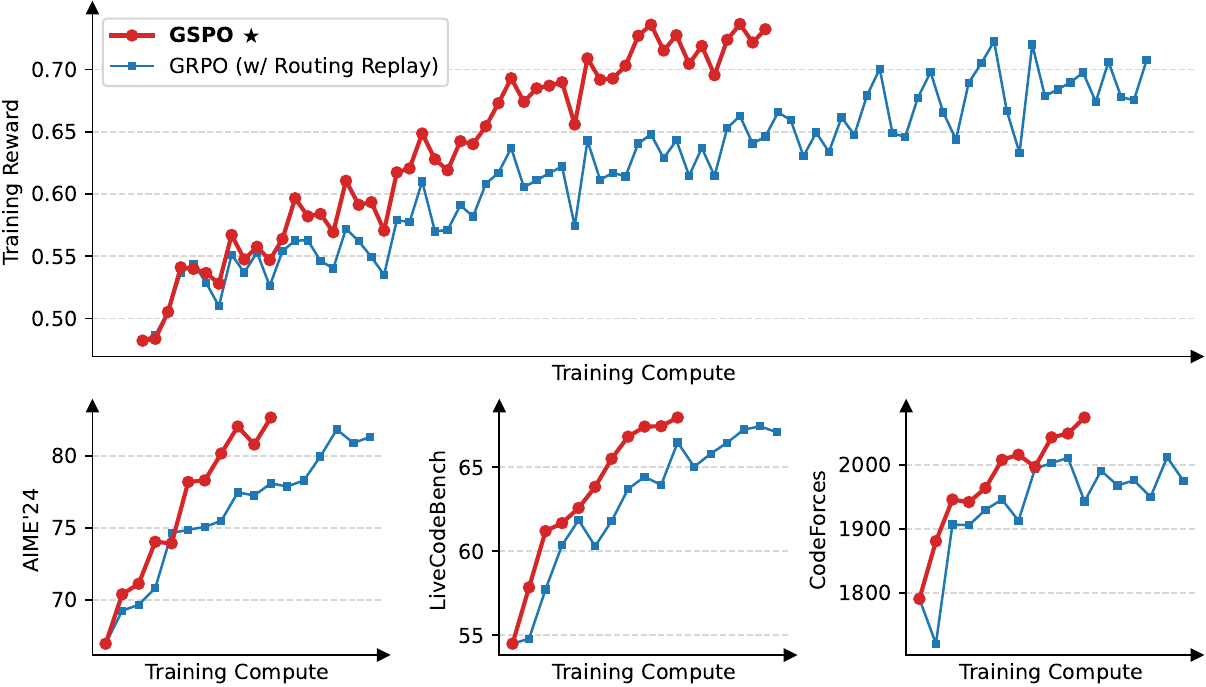}
    \caption{
        Training curves of a cold-start model fine-tuned from {Qwen3-30B-A3B-Base}.
        GSPO possesses remarkably higher training efficiency than GRPO.
    }
    \label{fig:results}
\end{figure}

Figure~\ref{fig:results} shows that the training with GSPO proceeds stably throughout.
We observe that \textbf{GSPO can deliver continuous performance improvement through increasing the training compute, regularly updating the query set, and extending the generation length}.
Moreover, GSPO also demonstrates superior training efficiency over GRPO, achieving better training accuracy and benchmark performance under the same training compute and consumed queries.
Finally, we have successfully applied GSPO to the RL training of the latest Qwen3 models, strongly proving the efficacy of GSPO in unleashing the power of RL scaling for large language models.

\subsection{Curious Observation on Clipping Fractions}

\begin{figure}[htbp]
    \centering
    \includegraphics[width=0.9\linewidth]{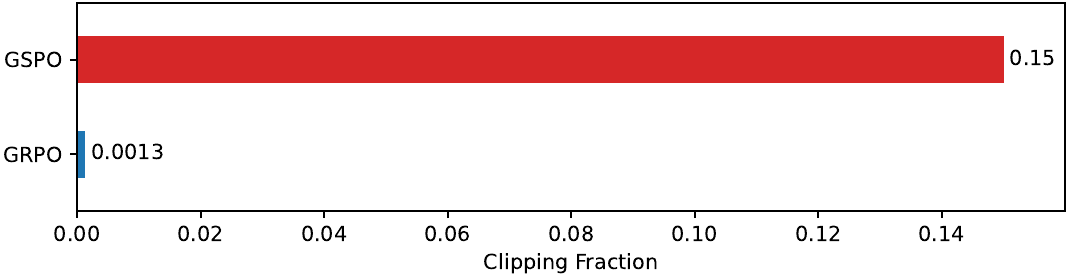}
    \caption{
        Average fractions of clipped tokens over the RL training of GSPO and GRPO.
    }
    \label{fig:clipping}
\end{figure}

A key distinction of GSPO compared to GRPO is its practice of clipping entire responses rather than individual tokens.
Particularly, as shown in Figure~\ref{fig:clipping}, we observe a difference of two orders of magnitude in the fractions of clipped tokens between GSPO and GRPO (while adjusting the clipping ranges does not alter the disparity in magnitude).
However, despite clipping significantly more tokens and consequently using fewer for training (or gradient estimation), GSPO still achieves higher training efficiency than GRPO.
This counter-intuitive finding — that clipping a much larger fraction of tokens leads to superior training efficiency — further indicates that GRPO's token-level gradient estimates are inherently noisy and inefficient for sample exploitation.
In contrast, GSPO's sequence-level approach provides a more reliable and effective learning signal.

\subsection{Benefit of GSPO for MoE Training}
\label{subsec:routing_replay}

\paragraph{Background}
Compared to the RL training of dense models, the sparse activation nature of MoE models introduces unique stability challenges.
In particular, we found that when adopting the GRPO algorithm, the \textit{expert-activation volatility} of MoE models can prevent RL training from converging properly.
To be specific, after one or more gradient updates, the experts activated for the same response can change significantly.
For example, with the 48-layer Qwen3-30B-A3B-Base model, after each RL gradient update and for the same rollout sample, there are roughly 10\% of the experts activated under the new policy $\pi_{\theta}$ that are different from those under the old policy $\pi_{\theta_\text{old}}$.
This phenomenon, which becomes more prominent in deeper MoE models, makes the token-level importance ratios $w_{i,t}(\theta) = \frac{ \pi_{\theta} (y_{i,t} | x, y_{i,<t}) }{ \pi_{\theta_\text{old}} (y_{i,t} | x,y_{i,<t})}$ fluctuate drastically and further invalidates them, as discussed in \S~\ref{sec:motivation} and~\ref{subsec:gradient}, consequently hindering the normal convergence of RL training.

\begin{figure}[htbp]
    \centering
    \includegraphics[width=\linewidth]{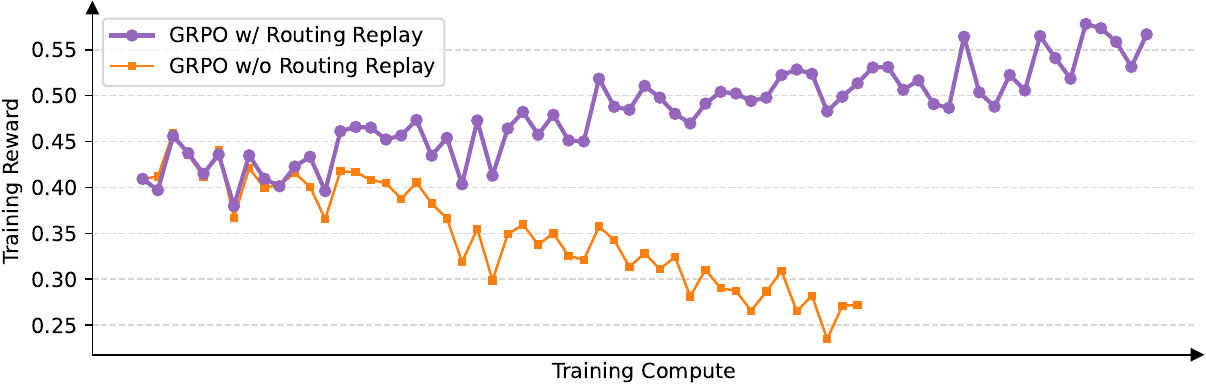}
    \caption{
        The Routing Replay strategy plays a critical role in the normal convergence of the GRPO training of MoE models.
    }
    \label{fig:routing_replay}
\end{figure}

\paragraph{Our Previous Approach}
To tackle this challenge, we previously employed the \textbf{Routing Replay} training strategy.
Specifically, we cache the activated experts in $\pi_{\theta_\text{old}}$ and ``replay'' these routing modes in $\pi_{\theta}$ when computing the importance ratios $w_{i,t}(\theta) = \frac{ \pi_{\theta} (y_{i,t} | x, y_{i,<t}) }{ \pi_{\theta_\text{old}} (y_{i,t} | x,y_{i,<t})}$.
In this way, for each token $y_{i,t}$, $\pi_{\theta} (y_{i,t} | x, y_{i,<t})$ and $\pi_{\theta_\text{old}} (y_{i,t} | x,y_{i,<t})$ share the same activated network, so that we can restore the stability of the token-level importance ratios and ensure optimization of the consistent activated network across gradient updates.
Figure~\ref{fig:routing_replay} demonstrates that Routing Replay serves as an essential technique in the normal convergence of the GRPO training of MoE models.

\paragraph{Benefit of GSPO}
Although Routing Replay enables the GRPO training of MoE models to converge properly, its practice of reusing routing modes incurs additional memory and communication overhead and can also limit the actual capacity of the MoE model.
In contrast, as shown in Figure~\ref{fig:results}, GSPO eliminates the dependency on Routing Replay and is fully capable of computing the importance ratios $s_i(\theta)$ conventionally, converging normally, and optimizing stably.
The key insight is that GSPO focuses only on the sequence likelihood (i.e., $\pi_{\theta} (y_i | x)$) and is not sensitive to the individual token likelihood (i.e., $\pi_{\theta} (y_{i,t} | x, y_{i,<t})$).
Since the MoE model always maintains its language modeling capability, the sequence likelihood will not fluctuate drastically.
In summary, GSPO fundamentally resolves the expert-activation volatility issue in MoE models, obviating the need for complex workarounds like Routing Replay.
This not only simplifies and stabilizes the training process but also allows the model to leverage its full capacity without artificial constraints.

\subsection{Benefit of GSPO for RL Infrastructure}

Given the precision discrepancies between training engines (e.g., Megatron) and inference engines (e.g., SGLang and vLLM), in practice, we typically use the training engine to recompute the likelihoods of sampled responses under the old policy $\pi_{\theta_\text{old}}$.
However, GSPO uses only sequence-level, rather than token-level, likelihoods for optimization, and intuitively, the former is much more tolerant of precision discrepancies.
Hence, GSPO makes it possible to directly use the likelihoods returned by the inference engine for optimization, thereby avoiding the need for recomputation with the training engine.
This can be especially beneficial in scenarios like partial rollout and multi-turn RL and in the training-inference disaggregated frameworks.

\section{Conclusion}

We propose Group Sequence Policy Optimization (GSPO), a new reinforcement learning algorithm for training large language models.
Following the basic principle of importance sampling, GSPO defines importance ratios based on sequence likelihood and performs sequence-level clipping, rewarding, and optimization.
GSPO demonstrates notably superior training stability, efficiency, and performance compared to GRPO and exhibits particular efficacy for the large-scale RL training of MoE models, laying the foundation for the exceptional improvements in the latest Qwen3 models.
With GSPO as a scalable algorithmic cornerstone, we will continue to scale RL and look forward to the resulting fundamental advances in intelligence.

\bibliographystyle{plainnat}
\bibliography{reference}

\end{document}